\documentclass[10pt,twocolumn,letterpaper]{article}

\usepackage[pagenumbers]{cvpr} 

\usepackage{graphicx}
\usepackage{amsmath}
\usepackage{amssymb}
\usepackage{booktabs}
\usepackage{verbatim}

\usepackage[pagebackref,breaklinks,colorlinks]{hyperref}

\usepackage{algorithm}
\usepackage{algpseudocode}
\usepackage{multirow}
\usepackage{hyperref}

\usepackage[capitalize]{cleveref}
\crefname{section}{Sec.}{Secs.}
\Crefname{section}{Section}{Sections}
\Crefname{table}{Table}{Tables}
\crefname{table}{Tab.}{Tabs.}

\def\cvprPaperID{3346} 
\def\confName{CVPR}
\def\confYear{2023}

\makeatletter
\g@addto@macro\@maketitle{
\vspace{-3em}
\begin{figure}[H]
\setlength{\linewidth}{\textwidth}
\setlength{\hsize}{\textwidth}
\centering
\includegraphics[trim={0cm, 0cm, 0cm, 0.0cm},clip,width=\linewidth]{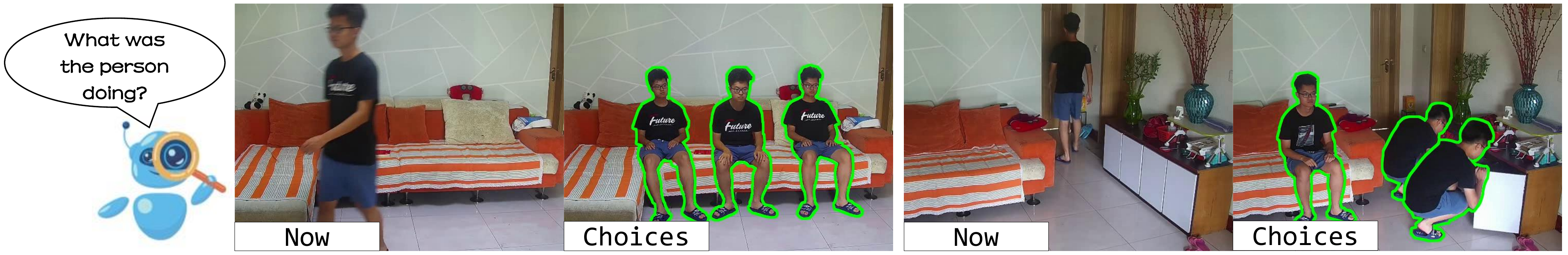}
\vspace{-6.5mm}
\caption{
{\textbf{Can you tell what the person was doing 3 seconds ago?} Inferring past human motions solely based on a single RGB image suffers from huge uncertainty. See the answers in \cref{fig:teaser2}.}
}
\label{fig:teaser1}
\end{figure}
}
\makeatother

\begin{document}

\title{What Happened 3 Seconds Ago? Inferring the Past with Thermal Imaging}

\author{Zitian Tang$^1$\quad Wenjie Ye$^1$\quad Wei-Chiu Ma$^2$\quad Hang Zhao$^{1,3}$ \vspace{8pt}\\
$^1$IIIS, Tsinghua University\quad
$^2$CSAIL, MIT\quad $^3$Shanghai Qi Zhi Institute
}
\maketitle

{\let\thefootnote\relax\footnote{Corresponding to: hangzhao@mail.tsinghua.edu.cn.}}

\begin{abstract}
   Inferring past human motion from RGB images is challenging due to the inherent uncertainty of the prediction problem. Thermal images, on the other hand, encode traces of past human-object interactions left in the environment via thermal radiation measurement. Based on this observation, we collect the first RGB-Thermal dataset for human motion analysis, dubbed Thermal-IM. Then we develop a three-stage neural network model for accurate past human pose estimation. Comprehensive experiments show that thermal cues significantly reduce the ambiguities of this task, and the proposed model achieves remarkable performance. The dataset is available at \href{https://github.com/ZitianTang/Thermal-IM}{https://github.com/ZitianTang/Thermal-IM}.

\end{abstract}

\section{Introduction}

    Imagine we have a robot assistant at home. When it comes to offering help, it may wonder what we did in the past. For instance, it wonders which cups were used, then cleans them. Or it can better predict our future actions once the past is known. But how can it know this? Consider the images in \cref{fig:teaser1}. Can you tell what happened 3 seconds ago? An image contains a wealth of information. The robot may extract geometric and semantic cues, infer the affordance of the scene, and imagine how humans would interact and fit in the environment. Therefore, in the left image, it can confidently deduce that the person was sitting on the couch; however, it is not sure where. Similarly, it can imagine many possibilities in the right image but cannot be certain. Indeed, given a single RGB image, the problem is inherently ill-posed.

\begin{figure*}
            \centering
            \includegraphics[width=1\linewidth]{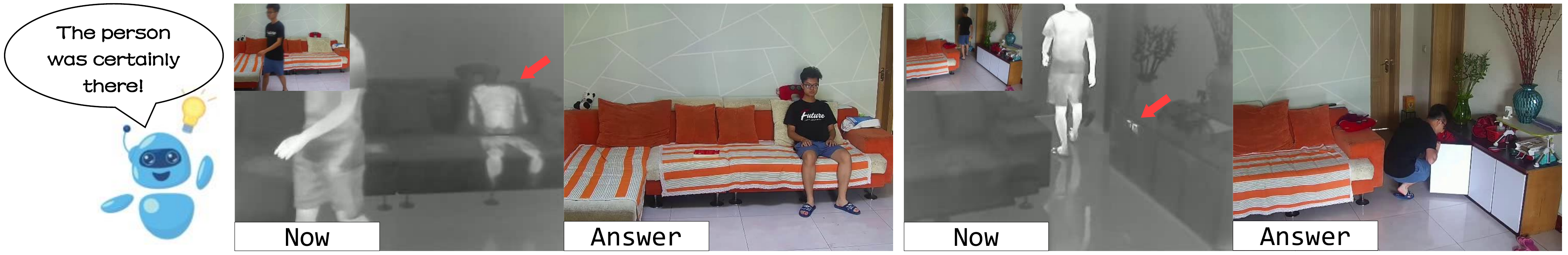}
            \caption{\textbf{Thermal images to the rescue:}
            Thermal images encode traces of past human-object interactions, which can help us infer past human behavior and understand objects' affordance. In this work, we focus on estimating human body poses a few seconds ago.}
            \label{fig:teaser2}
        \end{figure*}

    In this paper, we investigate the use of a novel sensor modality, thermal data, for past human behavior analysis. Thermal images are typically captured by infrared cameras, with their pixel values representing the temperature at the corresponding locations in the scene.
    As heat transfer occurs whenever there is contact or interaction between human bodies and their environment, thermal images serve as strong indicators of where and what has happened.
    Consider the thermal images in Fig. \ref{fig:teaser2}. With thermal images, we can instantly determine where the person was sitting.
    This is because the objects they contacted were heated, leaving behind bright marks. If a robot assistant is equipped with a thermal camera, it can more effectively infer the past and provide better assistance. Otherwise, we may need a camera installed in every room and keep them operational throughout the day.

    With these motivations in mind, we propose to develop a system that, given an indoor thermal image with a human in it, generates several possible poses of the person $N (N=3)$ seconds ago.
    To achieve this goal, we first collect a Thermal Indoor Motion dataset (Thermal-IM) composed of RGB, thermal, and depth videos of indoor human motion with estimated human poses. In each video, the actor performs various indoor movements (\eg, walking, sitting, kneeling) and interacts with different objects (\eg, couch, chair, cabinet, table) in a room.
    Then we design a novel, interpretable model for past human pose estimation. The model consists of three stages: the first stage proposes where the human might have been 3 seconds ago, leveraging the most discernible information in thermal images. The second stage infers what action the human was performing. Finally, the third stage synthesizes an exact pose.

    Experiments show that our method managed to generate plausible past poses based on the locations and shapes of thermal cues. These results are more accurate than the RGB-only counterparts, thanks to the reduced uncertainty of past human movements. Furthermore, our model automatically and implicitly discovers the correlation between thermal mark intensity and time.

    The contributions of this work are the following:
        \begin{itemize}
            \item We make the first attempt at a novel past human motion estimation task by exploiting thermal footprints.
            \item We construct the Thermal-IM dataset, which contains synchronized RGB-Thermal and RGB-Depth videos of indoor human motion.
            \item We propose an effective three-stage model to infer past human motion from thermal images.
        \end{itemize}

\section{Related Works}
    \paragraph{Thermal imaging in machine learning:}
        A thermal camera captures the far-infrared radiation emitted by any object (known as black body radiation), which is robust in varied illumination conditions. This property helps improve the performances of semantic segmentation and tracking systems significantly in urban scenes. Ha~\etal~\cite{Ha2017} releases the first RGB-Thermal image segmentation dataset and verifies the benefit of incorporating thermal images, especially in night-time scenes. Subsequently, plenty of datasets and models about semantic segmentation \cite{Zhou2022,Zhang2021,Deng2021,Sun2021,Lan2021,Vertens2020,Shivakumar2020,Sun2019} and tracking \cite{Li2016,Li2017,Li2019,Kristan2019,Li2022,Zhang2022,Zhang2022a,Luo2019,Zhang2021a,Wang2020,Zhang2019,Zhang2018,Jingchao2021} are proposed. These works propose various model structures to investigate the best way to fuse RGB and thermal features.

        There are a few works making use of other characteristics of thermal imaging. Based on the property that most glass is opaque to infrared light, Huo~\etal~\cite{Huo2022} recognizes glass based on RGB-Thermal image pairs. Their method significantly outperforms the RGB counterpart. As hand-object contact can leave apparent marks on objects, Brahmbhatt~\etal~\cite{Brahmbhatt2019} proposes a dataset recording contact maps for human grasps. They use a generative adversarial network model to predict how humans grasp a given object. Their results reveal various aspects affecting human grasping behavior.

        Our work is the first study on the relationship between thermal imaging and indoor human motion. Solving this task requires a deep understanding of how a thermal mark's location, shape, and intensity relate to human behavior.

    \paragraph{Human motion prediction:}

        Human motion prediction aims to predict the 2D or 3D future poses, given one's pose history. A wide range of techniques are used to tackle this task regardless of the scene context, such as graphical models \cite{Brand2000}, recurrent neural networks \cite{Fragkiadaki2015,Jain2016,Zhou2018,Martinez2017}, graph convolutional networks \cite{Zhong2022,Mao2019}, and temporal convolutional networks \cite{Hernandez2019,Li2018}. Moreover, \cite{Chao2017,Walker2017, Weng2019,Zhang2019a} consider pose history together with image context to predict future poses. However, these methods only concern a local patch around the human rather than the whole background scene.

        To predict 3D future human motion, Cao~\etal~\cite{caoHMP2020} proposes a method composed of three modules, GoalNet, PathNet, and PoseNet. Given an image and pose history, a VAE-based GoalNet predicts a few possible human torso positions in the future. Afterward, PathNet, an Hourglass model~\cite{Newell2016}, generates a route from the current human position to each predicted future one. Finally, Transformer-based PoseNet synthesizes a pose at each point along a route. It is worth noticing that the last two modules are deterministic, and the PoseNet is not provided with scene context. Wang~\cite{Wang2021} develops a GAN-based model to generate plausible future human motion in a given image. Their method comprises two stages. The first stage generates motion trajectories conditioned on the scene, while the second stage approximates the pose distribution given the scene and the trajectory.

        Our work focuses on inferring human motion in the past rather than the future. These two tasks are similar regardless of the direction of the time flow. Hence, the works above inspire our model design. While these works take a historical pose sequence into account, our work infers the past only according to a single frame and a static human pose in it.

\section{Thermal Indoor Motion dataset}
\label{sec:dataset}

        \begin{figure*}
            \centering
            \includegraphics[width=1\linewidth]{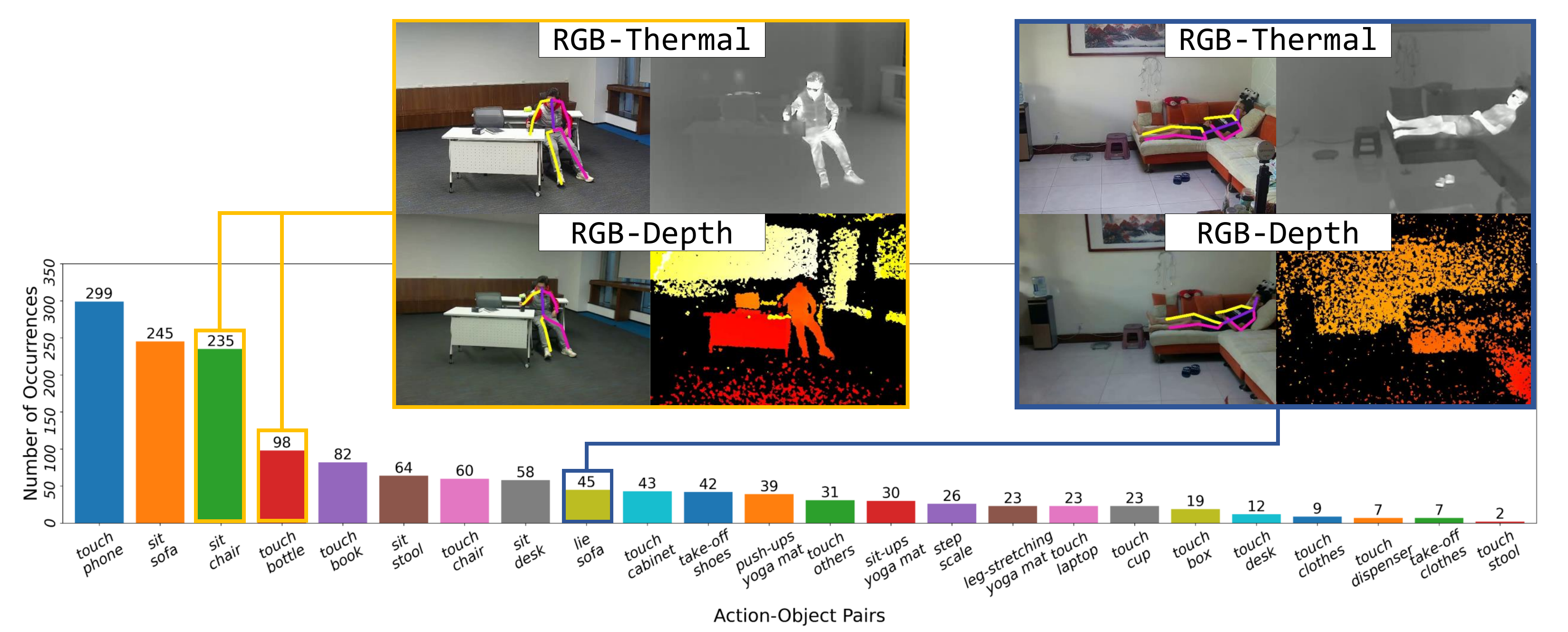}
            \caption{\textbf{Statistics of our Thermal Indoor Motion (Thermal-IM) dataset.} Our dataset comprises 783 synchronized RGB-Thermal and RGB- Depth video clips, with 24 types of human-object interactions. We also provide estimated 2D and 3D human poses in each frame.}
            \label{fig:dataset_stat}
        \end{figure*}

    Previous RGB-Thermal image datasets are mostly about urban scenes rather than indoor scenes. Moreover, none of them focus on human-scene interactions. This motivates us to collect the Thermal Indoor Motion (Thermal-IM) dataset. It contains synchronized RGB-Thermal and RGB-Depth videos with estimated human poses about a person moving and interacting with objects in indoor scenes.

    We collect the Thermal-IM dataset using an RGB-Thermal camera (Hikvision DS-2TD4237T-10) and an RGB-Depth camera (Intel RealSense L515). The resolution of the RGB-Thermal camera is $1080 \times 1920$ for the RGB channel and $288 \times 384$ for the thermal channel. That of the RGB-Depth camera is $480 \times 640$. These two cameras record videos simultaneously, and their extrinsic parameters are estimated.

    During data collection, an actor performs several preset actions, leaving cues in thermal images. In total, we collect 783 video clips, $\sim560$k frames in 15 FPS ($\sim10.4$ hours). $74\%$ of the videos involve one actor and two different rooms, which is the main part we use to develop our method. The rest is a held-out part for the generalization test in \cref{sec:analysis}, engaging one another actor or room. We record the videos from various viewing angles and rearrange the objects in the rooms to ensure scene diversity.

    We implement a pose estimation pipeline to derive smooth and accurate 3D pose sequences in the videos. Details of the pose estimation process are in Appendix. We manually annotate the start and end time of each human-object interaction in the videos. There are 24 different types of action-object pairs present in the dataset. Statistical details and examples are in \cref{fig:dataset_stat}.

    Although 3D poses and depth point clouds are available in the dataset, we concentrate on a 2D version of the proposed task - inferring 2D poses in the image space. Therefore, we obtain 2D poses by projecting the 3D ones to the image plane of the RGB-Thermal camera to conduct our work. Note that the actor is usually stationary, in which case motion inference is trivial. We filter out the clips where the average displacement per joint is less than 45 pixels in 3 seconds. The remaining $110$k frames serve as the data for our proposed task.

\section{Method}
    This work aims to infer what a person in a thermal image was doing $N$ seconds ago. We set $N=3$ since we empirically find that it takes at most 3 seconds for a person to complete an action. If $N$ is too small, one can infer past poses directly from current poses without any context. On the other hand, if $N$ gets larger, the thermal cues may disappear, and the uncertainty of the past increases.

    Due to the inherent uncertainty of human motion, our model makes stochastic predictions, \ie, $M$ possible 3s-ago poses of the person. Thermal images provide plenty of cues telling where the people were and how they interacted with the environment. It is challenging for a model to understand this information and make plausible inferences.

    Formally, given a thermal image $I\in\mathbb{R}^{H\times W}$, the goal is to generate $M$ 2D poses of the person 3 seconds ago, denoted by $q_{1:M}\in\mathbb{R}^{M\times J\times 2}$. Here $(H,W)$ is the size of the images, and $J$ is the number of joints in a human pose. We also provide the current pose $p\in\mathbb{R}^{J\times 2}$ of the person in the image so that a model can focus on inferring the past instead of struggling to recognize the person first.

        \begin{figure*}
            \centering
            \includegraphics[width=0.7\linewidth]{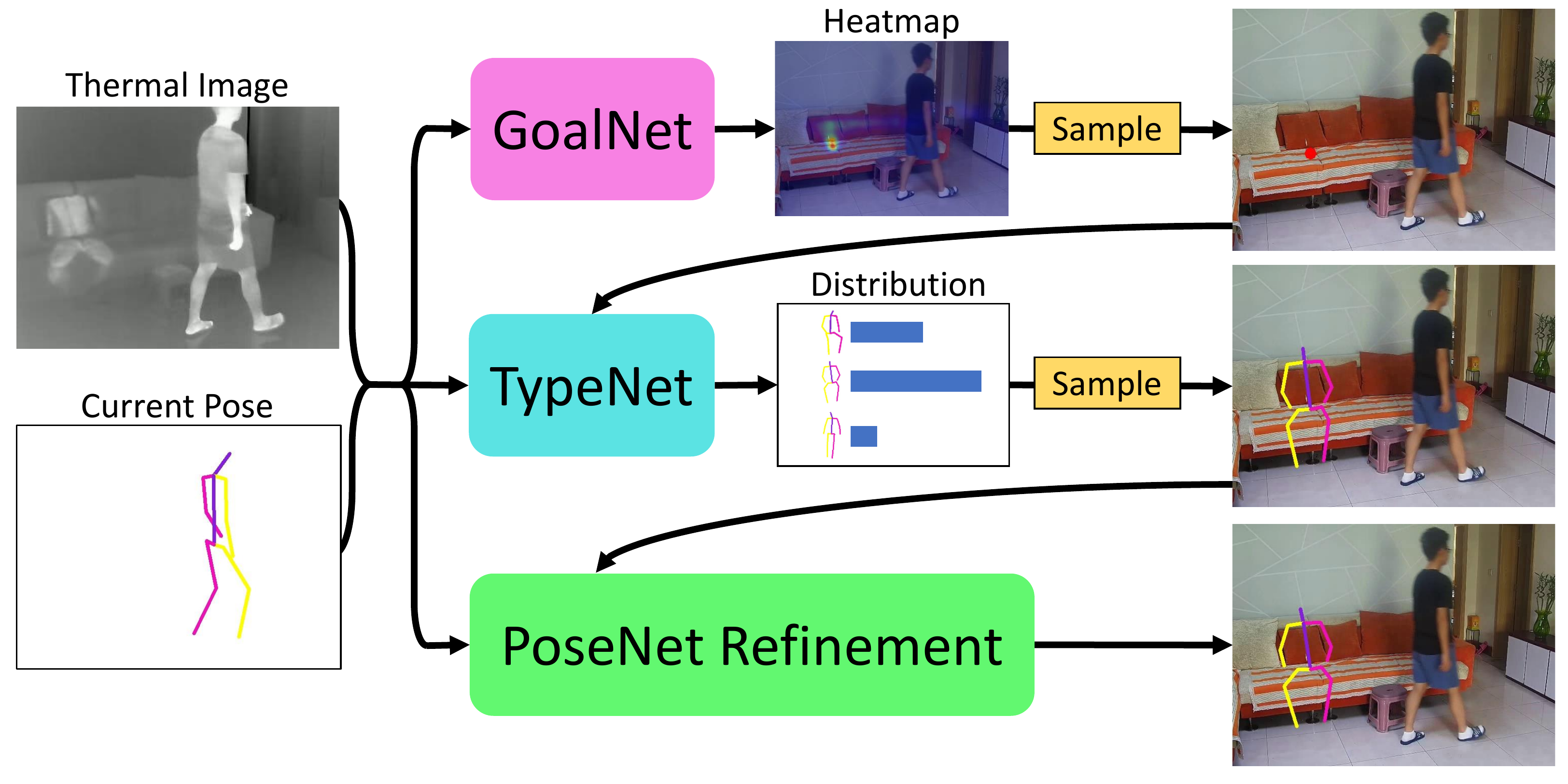}
            \caption{\textbf{Overview of our pipeline.} Given a thermal image and an estimated human pose as input, GoalNet first predicts the distribution of the person's location 3 seconds ago and samples one from it. Then, TypeNet predicts pose type distribution and samples one pose from them. Finally, PoseNet refines the pose to match the input observation better. The RGB images are for visualization purposes only.}
            \label{fig:pipeline}
        \end{figure*}

    We propose a three-stage framework illustrated in \cref{fig:pipeline} to tackle this task. It has three components, GoalNet, TypeNet, and PoseNet. In the first stage, GoalNet proposes possible positions where the human was 3 seconds ago. Next, TypeNet assigns a possible pose type (sitting, standing, walking, \etc.) at each proposed position. Finally, PoseNet synthesizes a pose of the assigned type at each proposed position. The rest of this section will discuss the motivation and details of our method.

    \subsection{GoalNet}
        When a thermal image like \cref{fig:teaser2} is shown to humans, one can intuitively figure out where the actor used to be by recognizing the bright marks on the objects. Consequently, one can confidently tell that the actor was around the bright mark or on a path connecting that position with the current position 3 seconds ago. Therefore, our first stage GoalNet $\mathcal{G}$ is designed to capture the distribution of human position.

        Let $r\in\mathbb{R}^{2}$ denote the torso joint position of a pose $q\in\mathbb{R}^{J\times2}$ that we want to generate. We sample $r$ from a distribution $P(r)\in\mathbb{R}^{H\times W}$, where
        \begin{align}
            P(r)&=\mathcal{G}(I,H_p)
        \end{align}
        is predicted by GoalNet based on the image and current human pose. Here we use $H_x\in\mathbb{R}^{L\times H\times W}$ to denote the heatmap representation of a series of 2D positions $x\in\mathbb{R}^{L\times2}$. Thus $H_p$ here is in shape $J\times H\times W$.

        We use an Hourglass model \cite{Newell2016} as the architecture of GoalNet. This model can not only capture local thermal cues but also consider the context information to generate a plausible position distribution in the image space.

    \subsection{TypeNet}
        After the torso position is specified, the next step is to generate a pose at that location. Instead of drawing a human pose directly, one may first speculate what action the character was doing there, such as if the one was sitting or standing and if the one was facing to the left or right. Moreover, the possible human poses are diverse even at a particular position, \eg, it is plausible to stand to both the left and right in some circumstances. Therefore, we need first specify a \emph{pose type} (action) at each position before synthesizing an explicit pose.

        To derive pose labels, we cluster all the poses in the training set into several groups, and each group corresponds to a pose type. In practice, we align the torso joints of all poses, represent a $J$-joint pose as a $2(J-1)$-dimensional vector, and apply the K-Means algorithm in Euclidian space to form 200 clusters.

        The second stage TypeNet $\mathcal{T}$ then gives a distribution $P(z)\in\mathbb{R}^{200}$ over all pose types at the proposed position $r$ according to the inputs, where $z$ denotes a pose type index. Formally,
        \begin{align}
            P(z)&=\mathcal{T}(I,H_p,H_{r}).
        \end{align}
        As a typical image classification task, ResNet18 \cite{He2016} serves as the backbone of TypeNet. We can sample a pose type from the distribution $P(z)$ as the actor's action that we infer.

    \subsection{PoseNet}
        The final step is synthesizing a human pose of type $z$ at location $r$. At this step, the detailed information in the image determines the pose's size and the joints' accurate positions. We develop PoseNet $\mathcal{P}$ to infer the pose while being aware of this information.

        PoseNet is also an Hourglass model like GoalNet. It gives a heatmap $P(q)\in \mathbb{R} ^ {(J-1) \times H \times W}$ for all joints of $q$ except the torso joint. Instead of feeding the pose type index $z$ into PoseNet, we paint the $z$-th cluster's center pose at position $r$ as input. Hence, PoseNet is refining a given pose rather than generating a pose from scratch. Formally, we have
        \begin{equation}
            P(\tilde q)=\mathcal{P}(I,H_p,H_{r},H_{C_{z}+r}) ,
        \end{equation}
        \begin{equation}
            \forall 1\leq j\leq J-1,\;\tilde q_{j}=\mathop{\arg\max}\limits_{x,y}P(\tilde q_{j}=(x,y)),
        \end{equation}
        \begin{equation}
            q=\left[\tilde q, r\right],
        \end{equation}
        where $\tilde q$ denotes a human pose without the torso joint and $C_z$ is the $z$-th pose cluster center.

    \subsection{Learning}
        We split the main part of the dataset mentioned in \cref{sec:dataset} into training, validation, and test sets in terms of video clips and train our model with the training set.

        The three modules are trained separately, using the ground truth in the last step as input and supervised by the labels in the current step. As the predictions of all modules are probability distributions, we utilize Cross Entropy Loss ($\mathcal {L}_\text{CE}$) as their training objectives. Let $\hat r$ and $\hat q$ be the ground truth torso position and human pose 3 seconds ago, and let $\hat z$ be the pose type of $\hat q$. The training losses for GoalNet $\mathcal{G}$, TypeNet $\mathcal{T}$, and PoseNet $\mathcal{P}$ are
        \begin{equation}
            \mathcal{L_{G}} = \mathcal {L}_\text{CE}\left(\mathcal{G}(I,H_p), \hat r\right),
        \end{equation}
        \begin{equation}
            \mathcal{L_{T}} = \mathcal {L}_\text{CE}\left(\mathcal{T}(I,H_p, H_{\hat r}), \hat z\right),
        \end{equation}
        \begin{equation}
            \mathcal{L_{P}} = \sum_{j=1}^{J-1} \mathcal {L}_\text{CE} \left(\mathcal{P}_j(I,H_p, H_{\hat r}, H_{C_{\hat z} + \hat r}), \hat q_j\right).
        \end{equation}

    \subsection{Inference}
        This task requires a model to give $M$ possible answers for each test sample. To do this, we sample $M$ torso positions $r$ at the GoalNet stage and then run TypeNet and PoseNet once for each sampled position.

        In TypeNet, rather than sampling among all pose types, we find that top-$k$ sampling with $k=5$ leads to the best performance. That is, we sample the pose type from the five types with the highest probabilities given by TypeNet. At the PoseNet stage, the position with the highest weight in each joint's heatmap is picked as the final prediction.

\section{Experiments}
    In this section, we first evaluate the effectiveness of our approach on the Thermal-IM dataset.
    Then we investigate the importance of different modalities for inferring past human behavior.
    Finally, we comprehensively study the characteristic of our model.

        \begin{figure}
            \centering
            \includegraphics[width=1\linewidth]{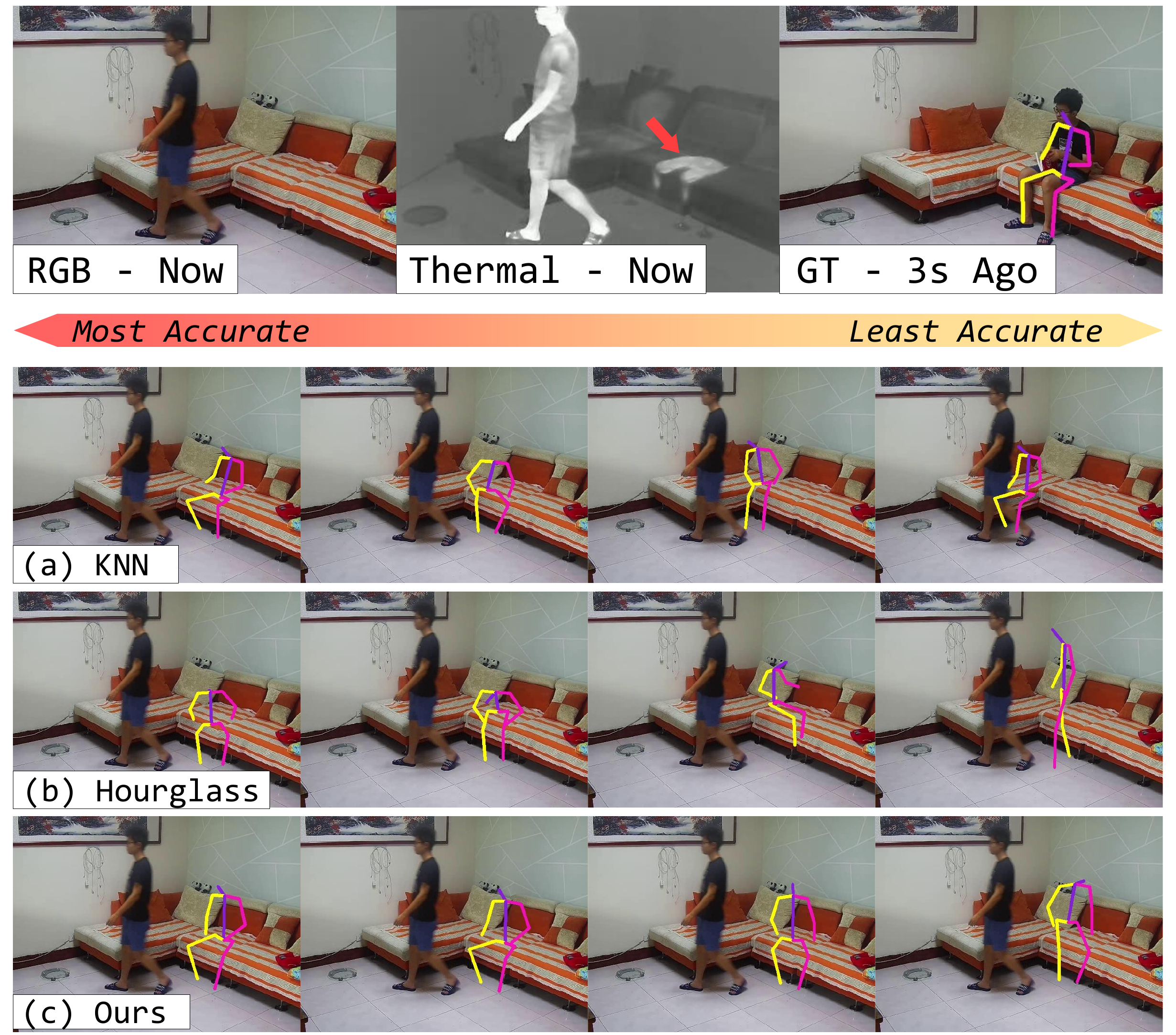}
            \caption{\textbf{Comparison against baselines.} We sort the predictions of each approach based on MPJPE and show the 1st, 5th, 10th, and 20th poses from left to right.
            KNN fails to infer where the person was. Hourglass is able to locate where the person was, but their predictions do not comply with the affordance of the scene. Our method, in contrast, produces reasonable and accurate estimates.}
            \label{fig:res1}
        \end{figure}

    \subsection{Evaluation metrics}
    \label{sec:metrics}

        \paragraph{MPJPE:} We calculate the Mean Per Joint Position Error (MPJPE) \cite{Ionescu2014} between the generated poses and the ground truth to evaluate their similarity. Specifically, MPJPE is the average Euclidian distance in the number of pixels from each joint to its corresponding answer. Due to the uncertainty of the past, 30 poses are generated for each test sample by each model. We report the average MPJPE of the top-1/3/5 ones closest to the ground truth.

        \paragraph{Negative log-likelihood (NLL):}
        As our modules yield probability distributions, we can determine the probability of each pose. To quantify the accuracy of our model, we utilize the NLL of the actual poses as a metric. We report this metric for all methods that support likelihood estimation.

        \paragraph{Semantic score:} Note that one can randomly synthesize diverse poses ignoring the scene context to achieve low top-$k$ MPJPE. However, such poses may be implausible in the scene. We use semantic score \cite{Li2019a} to measure how many generated poses are plausible in the given contexts. Specifically, we construct a dataset containing RGB images with plausible and implausible poses based on Thermal-IM (including the held-out part) and train a binary classifier to distinguish them. Plausible poses are the 3s-ago poses, and implausible poses are derived by randomly replacing, shifting, and perturbing the plausible ones. The classifier achieves a test accuracy of $85.77\%$. The semantic score for a method is defined as the ratio of generated poses recognized as plausible by the classifier. Examples of training data and implementation details are in Appendix.

    \subsection{Baselines}

        \paragraph{KNN:}
        We first construct a pool of current-past pose pairs from the training set. Since the adjacent video frames are alike, we sample one frame every 15 frames. Next, given a test human pose, we leverage K-Nearest Neighbor (KNN) to retrieve 30 closet samples from the pool. Finally, we treat the corresponding past pose as the results.

        \paragraph{Hourglass:}
        We adapt the state-of-the-art 2D pose estimation model \cite{Newell2016} as our second baseline. Given input observation(s), we first predict a distribution map for each joint of the \emph{past} pose.
        Since independently sampling each joint may result in unrealistic poses, we then exploit the human poses from the training set and evaluate their likelihood with the predicted distribution. This ensures that the estimated poses are always realistic.
        Finally, we select 30 poses with the highest likelihood.
        In practice, we consider 1/200 poses from the training set.

    \subsection{Evaluation results}

        \begin{table}
          \centering
          \begin{tabular}{cccccc}
            \toprule
            \multirow{2}{*}{Method} & \multicolumn{3}{c}{MPJPE} & \multirow{2}{*}{NLL} & Semantic\\
            \cmidrule(r){2-4}
            & Top 1 & Top 3 & Top 5 & & Score($\%$)\\
            \midrule
            KNN & 19.26  & 24.53 & 28.44 & N/A & 61.94\\
            Hourglass & 23.80 & 27.99 & 31.03 & 136.23 & 67.05\\
            \midrule
            Ours & \textbf{18.33} & \textbf{22.25} & \textbf{25.25} & \textbf{103.75} & \textbf{82.11} \\
            \bottomrule
          \end{tabular}
          \caption{\textbf{Evaluation results of our model and baselines.} Our model outperforms all the baselines in all metrics.}
          \label{tab:main_results}
        \end{table}

        \begin{table}
          \centering
          \begin{tabular}{cccccc}
            \toprule
            \multirow{2}{*}{Input} & \multicolumn{3}{c}{MPJPE} & \multirow{2}{*}{NLL} & Semantic\\
            \cmidrule(r){2-4}
            & Top 1 & Top 3 & Top 5 & & Score($\%$)\\
            \midrule
             RGB & 22.06 & 27.21 & 31.12 & 105.03 & \textbf{87.56} \\
            Thermal & \textbf{18.33} & \textbf{22.25} & \textbf{25.25} & \textbf{103.75} & 82.11 \\
             RGB-T & 19.23 & 23.52 & 26.76 & \textbf{103.75} & 85.46 \\
             T w/o pose & 19.62 & 24.00 & 27.27 & 104.38 & 80.55 \\
            \bottomrule
          \end{tabular}
          \caption{\textbf{Ablation study on model input.} The thermal model achieves the best MPJPE and NLL, while the RGB model has the highest semantic score. The RGB-T model access both modalities but does not provide a better performance in any metric. Once the current pose is not provided, the thermal model can still achieve competitive results.}
          \label{tab:ablation_modality}
        \end{table}

        As shown in \cref{tab:main_results}, our method outperforms the baselines significantly across all metrics. It is able to recover plausible human poses 3 seconds ago accurately. Some qualitative results are shown in \cref{fig:res1}.
        In the thermal image, the bright mark implies that the person was sitting on the sofa. Our method observes this and synthesizes several poses sitting or getting up from there. In contrast, KNN either retrieves poses sitting in other places or gives implausible answers - a pose sitting on nothing. As for Hourglass, although it succeeds in locating the place where the person was sitting, the estimated poses do not comply with the affordance of the sofa.

    \subsection{Ablation studies}
        We first investigate the importance of different modalities for inferring the past. Then we study whether the availability of the current human pose will affect the model performance. We refer the readers to Appendix for ablation on the modules.

            \begin{figure}
                \centering \includegraphics[width=1\linewidth]{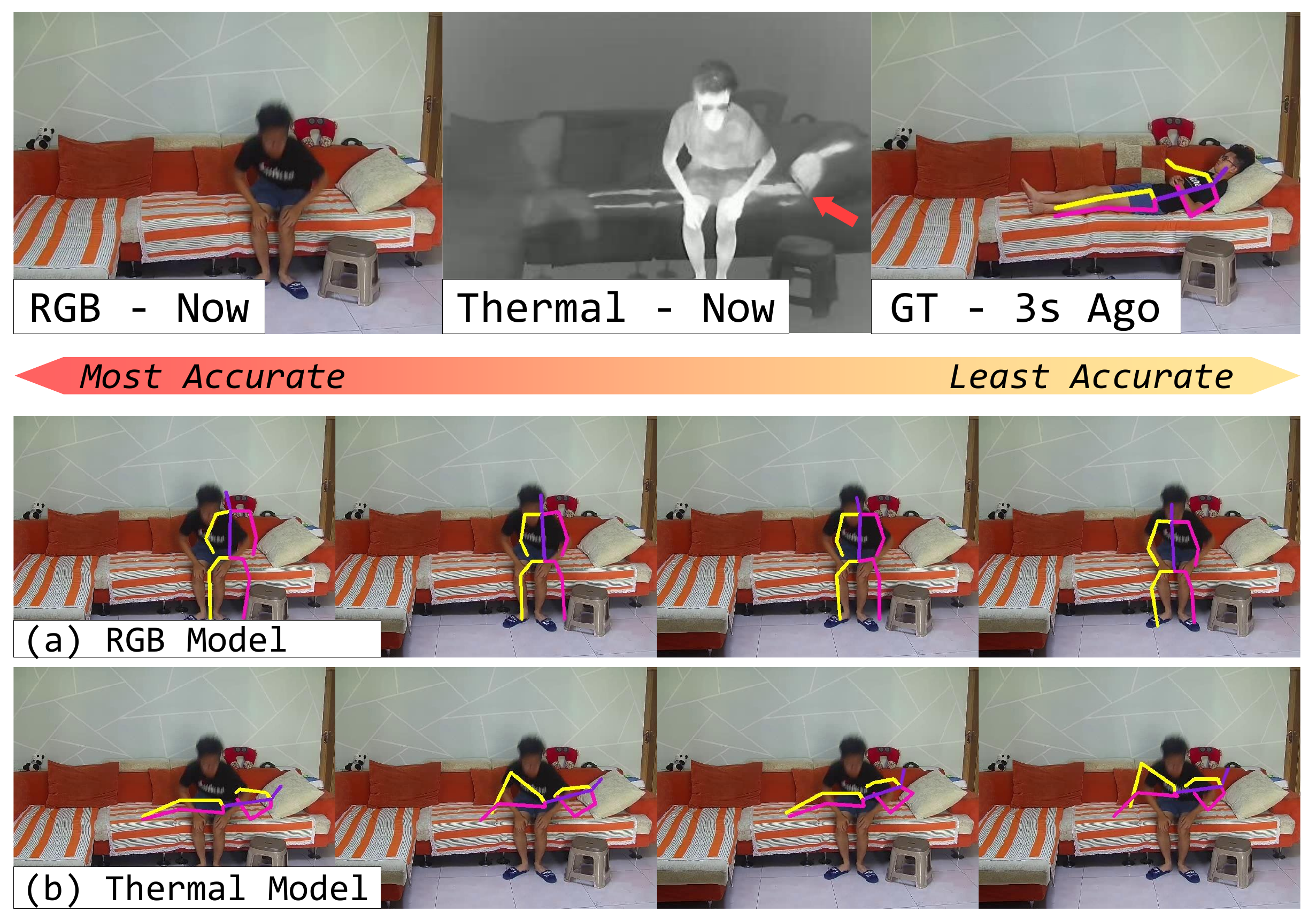}
                \caption{\textbf{Importance of thermal imaging.} In this example, it is hard to infer the person's past action through the RGB image. With the thermal image, however, one can easily and reliably infer that the person was lying on the couch.}
                \label{fig:res2}
            \end{figure}

            \begin{figure}
                \centering \includegraphics[width=1\linewidth]{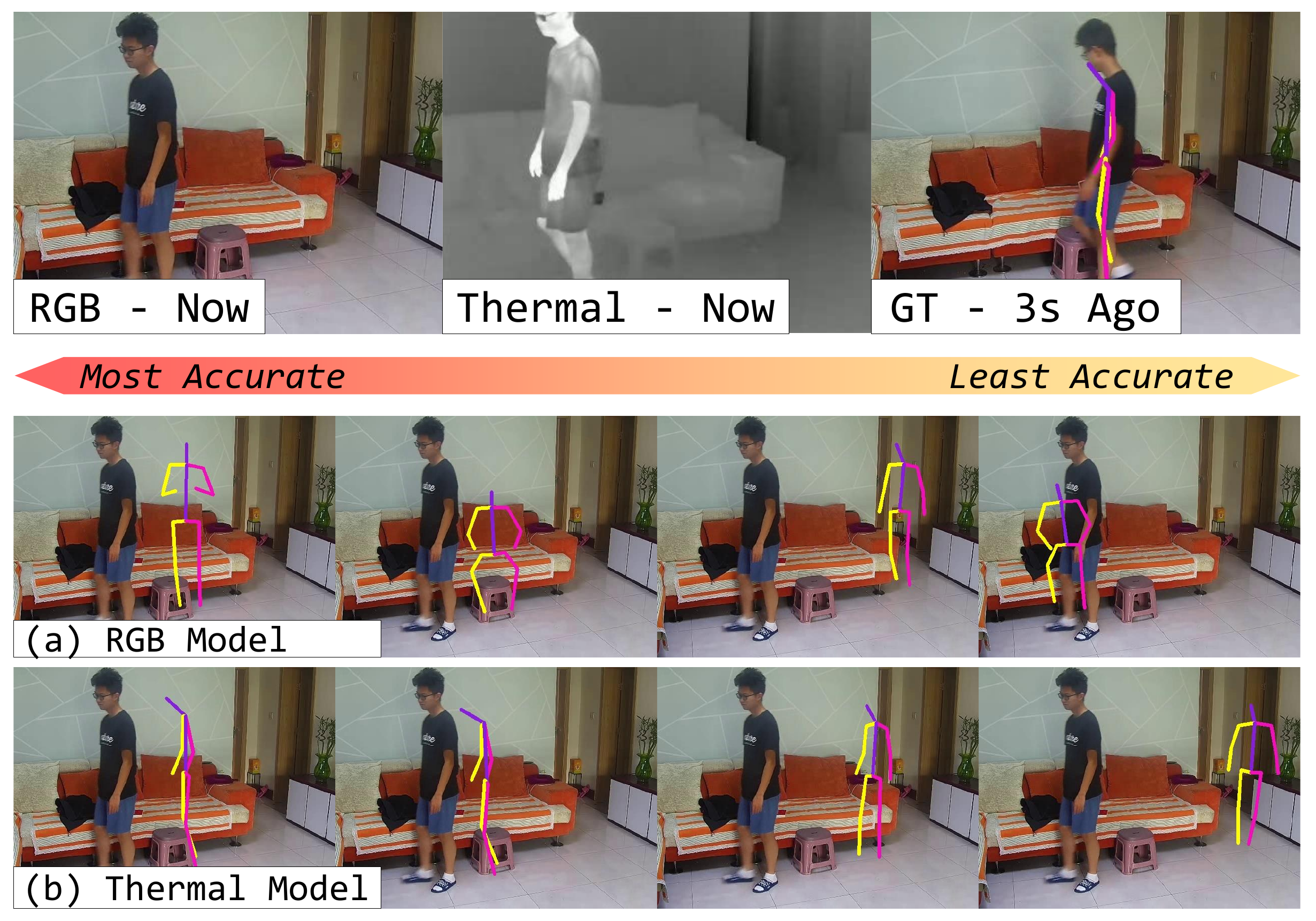}
                \caption{\textbf{Ablation study on input modality.} In this example, the person did not touch any object. While the thermal model is capable of inferring this reliably, the RGB model fails --- it predicts sitting poses incorrectly.}
                \label{fig:res4}
            \end{figure}

        \paragraph{Importance of different modalities:}
        As shown in \cref{tab:ablation_modality}, our model performs best on MPJPE and NLL when taking thermal images as input.
        However, the semantic score is higher when RGB images are included.
        We conjecture this is because the details in the scene are more apparent in the RGB domain.

        We show two qualitative comparisons in \cref{fig:res2} and \cref{fig:res4}.
        The horizontal thermal mark on the sofa (see \cref{fig:res2}) implies that the person was lying there.
        With thermal images, the model can infer various lying poses. Its RGB counterpart, however, cannot notice this and only synthesizes sitting poses.
        As for \cref{fig:res4}, there is no thermal cue on any object, indicating that the person did not touch anything in the short past.
        The thermal model thus only generates walking poses.
        In contrast, the RGB model fails to capture the difference and predicts multiple sitting poses.

        Additionally, although the RGB-Thermal model benefits from richer information, its performance falls between that of the RGB and thermal models. This observation suggests that early fusion methods, such as concatenation at the input level, fail to capture cross-modal interactions effectively. Further research is needed to develop more effective fusion methods that can capitalize on the complementary nature of RGB and thermal modalities.

        \paragraph{Current pose as input:} {To infer the past, it is crucial for a model to know the human pose in the current frame. Once the model is provided with the current pose, it does not need to implicitly learn to recognize the human. However, one may have trouble estimating the current pose in practice. To tackle this issue, we train a model that does not require current poses. As shown in \cref{tab:ablation_modality}, the performance degrades a bit but is still competitive. If the human pose is unavailable in practice, our method without input body pose can serve as an effective alternative.}

    \subsection{Analysis}
    \label{sec:analysis}
        \begin{figure}
            \centering
            \includegraphics[width=1\linewidth]{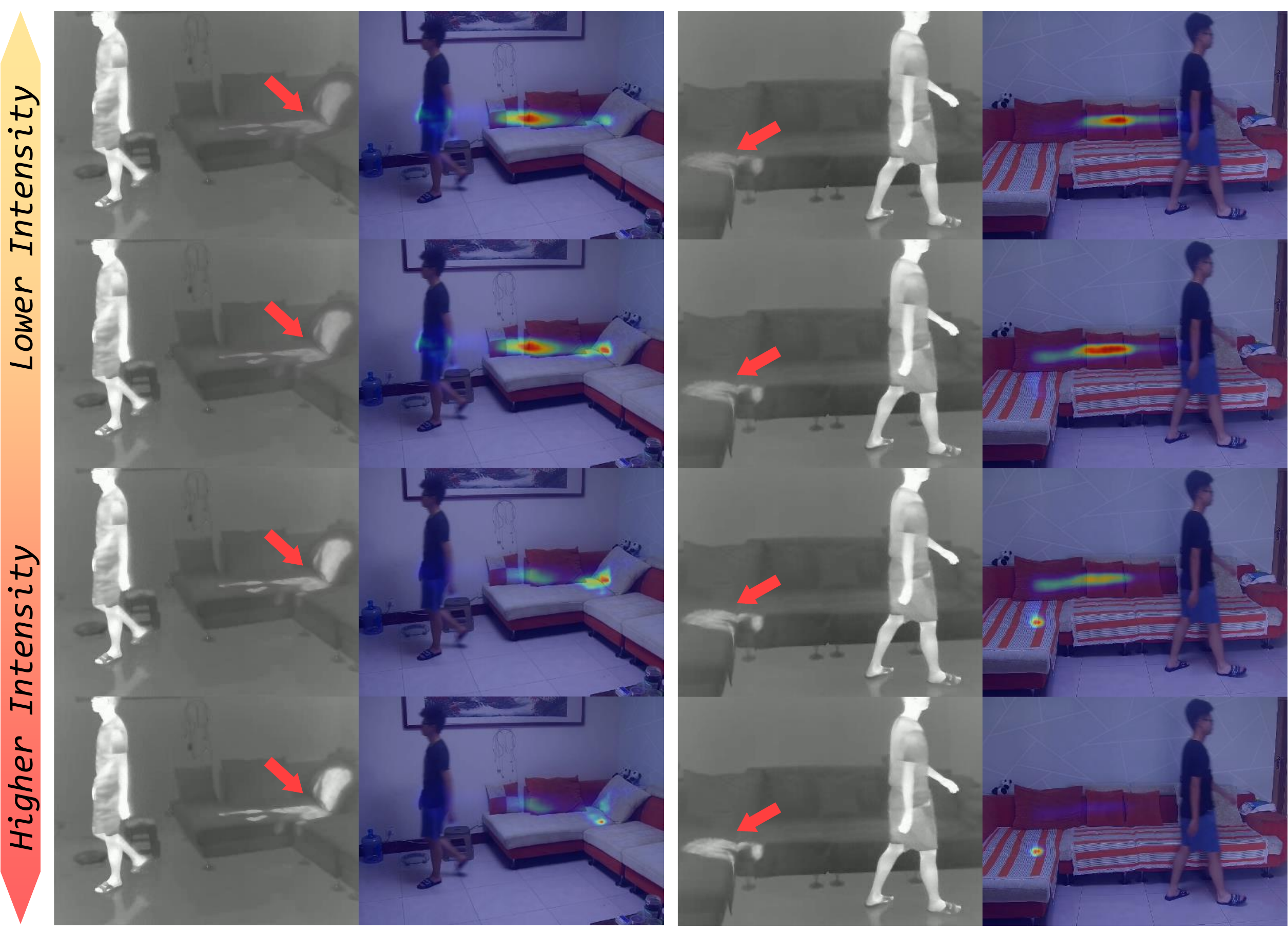}
            \caption{
            \textbf{Effect of thermal intensity on GoalNet predictions.}             
            The intensity indicates how long the time has passed since the last interaction --- the larger the intensity, the shorter amount of time.
            As it increases, the inferred distribution of the character's 3-second-ago position gets closer to the thermal mark.}
            \label{fig:intensity}
        \end{figure}

        \begin{figure*}
            \centering
            \includegraphics[width=1\linewidth]{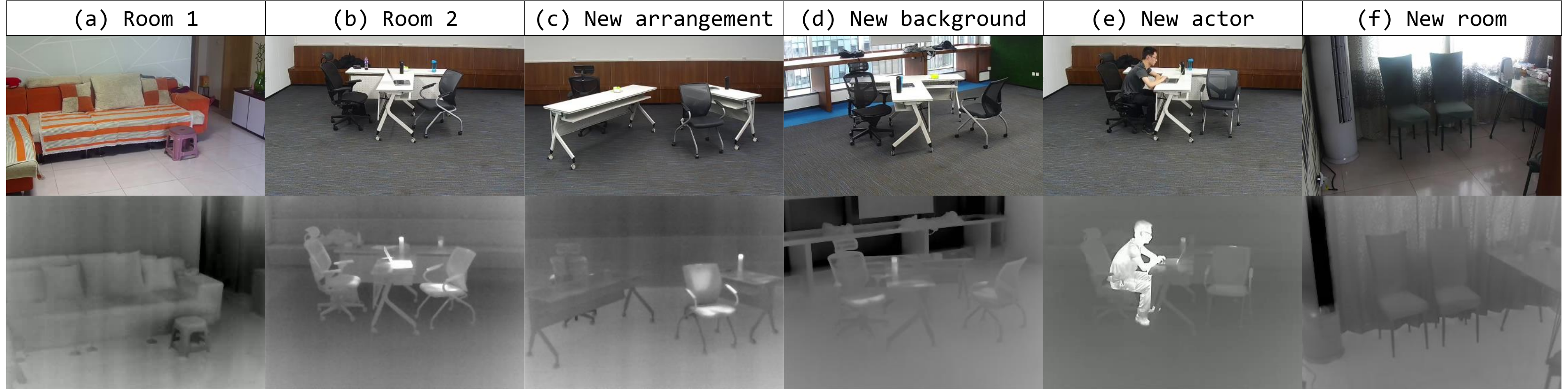}
            \caption{
            \textbf{Generalization to new environment:} 
            Our training data is collected from two rooms: (a) Room 1 and (b) Room 2. 
            To verify whether our model can generalize to new environment without overfitting, during evaluation we (c) rearrange the object layouts, (d) switch the background, (e) replace the actor, and even (f) test on a complete different room.}
            \label{fig:genralization}
        \end{figure*}

        \begin{table*}
          \centering
          \begin{tabular}{ccccccc}
            \toprule
            \multirow{2}{*}{Changed Factor} & \multirow{2}{*}{Modality} & \multicolumn{3}{c}{MPJPE} & \multirow{2}{*}{NLL} & \multirow{2}{*}{Semantic Score($\%$)}\\
            \cmidrule(r){3-5}
            & & Top 1 & Top 3 & Top 5 & & \\
            \midrule
            \multirow{2}{*}{Arrangement} & RGB & 21.27 & 26.38 & 30.42 & 107.10 & \textbf{93.69}\\
            & Thermal & \textbf{20.41} & \textbf{25.10} & \textbf{28.36} & \textbf{105.37} & 89.56\\
            \midrule
            \multirow{2}{*}{Background} & RGB & 25.07 & 30.02 & 33.47 & 111.67 & \textbf{83.80} \\
             & Thermal & \textbf{19.85} & \textbf{24.24} & \textbf{27.83} & \textbf{107.82} & 81.49 \\
            \midrule
            \multirow{2}{*}{Actor} & RGB & \textbf{24.37} & 29.20 & 32.77 & 114.87 & \textbf{91.21}\\
            & Thermal & 24.60 & \textbf{28.92} & \textbf{31.98} &  \textbf{114.26} & 81.33\\
            \midrule
            \multirow{2}{*}{Room} & RGB & 35.05 & 42.00 & 47.11 & 121.14 & 19.55\\		
            & Thermal & \textbf{23.05} & \textbf{27.59} & \textbf{31.16} & \textbf{112.84} & \textbf{36.88}\\
            \bottomrule
          \end{tabular}
          \caption{\textbf{Generalization test results.} In most cases, the thermal model provides more accurate predictions, while the RGB model achieves higher semantic scores when the room is not changed. Moreover, the thermal model greatly outperforms the RGB model when introducing a new background or room. This certifies that our thermal model is more robust to environmental appearance.}
          \label{tab:generalization}
        \end{table*}

        \paragraph{Effect of thermal intensity:} After a human-object interaction, the thermal mark left on the object gradually gets dimmer and finally vanishes. Thus the intensity of the mark reveals information about when the interaction happened. To see whether our model learns about this knowledge, we manually modify the mark brightness in a given image and then examine how our model's prediction changes.

        \cref{fig:intensity} shows the varied heatmap predictions of GoalNet, representing the distributions of the person's position 3 seconds ago. When the mark on the sofa is bright, the heatmap density tends to converge at the mark's position. On the contrary, it is close to the person when the mark is dim. This result coincides with our intuition. A mark is bright only if the interaction was just over in the past few seconds; therefore, it is more likely that a person was there 3 seconds ago. Conversely, if a mark is dim, the person probably had already left there 3 seconds ago. This experiment suggests that our model understands the time information contained in thermal mark intensity.

        \paragraph{Generalization:}
        The videos in the Thermal-IM training set only involve one actor and two different rooms. It is critical to investigate whether a model trained on it generalizes well when the actor and environment are changed. To this end, we use the held-out part of the Thermal-IM dataset mentioned in \cref{sec:dataset} to conduct generalization experiments. This part of data involves several factors changed from the training set, including the arrangement of objects, the background, the actor, and the room. These changes are illustrated in \cref{fig:genralization}.

        We test our RGB and thermal models over these four cases to examine how these changes influence them. \cref{tab:generalization} shows the results. Although the model performances deteriorate compared to \cref{tab:ablation_modality}, the thermal model is still the best in MPJPE and NLL when changing the arrangement, background, or room. Particularly, when a new background or room is involved, the performance decrease of the thermal model is much smaller than that of its RGB counterpart. We hypothesize that reason is that when seeing new objects and backgrounds, the RGB model tries to identify things humans can interact with; instead, the thermal model infers the past by finding the thermal marks in certain shapes without considering object identification. The latter mechanism is more robust to the appearance changes of the environment.

        \paragraph{Limitations:} In \cref{tab:generalization}, we observe performance degradation in both RGB and thermal models when a new actor is involved. We ascribe this to the new actor's different stature and behavioral habits from the one in the dataset. The personal habits introduce new actions the model has never seen, and the different statures indicate different sizes of human poses. However, our model architecture, most notably TypeNet, limits the prediction of poses that appear in the training set. We expect future work on the model design to perform better on actor generalization.

\section{Conclusions}
    In this work, we propose to infer past human pose by leveraging thermal imaging. We collect the Thermal-IM dataset containing RGB-Thermal and RGB-Depth videos about indoor human motion with estimated poses. Based on this dataset, a three-stage method is developed to tackle the proposed task. We show that inference of the past becomes an easier task with thermal images compared to RGB ones. The experiments demonstrate not only our model's capability of understanding past human location and action but also its awareness of the correlation between thermal mark intensity and time. Some aspects of this task remain to be explored, such as how to effectively fuse RGB and thermal modalities to use their information jointly.

    \paragraph{Acknowledgement} WCM is partially funded by a Siebel scholarship and the MIT-IBM Watson AI Lab.

{\small
\bibliographystyle{ieee_fullname}
\bibliography{ref}

\begin{thebibliography}{10}\itemsep=-1pt

\bibitem{Brahmbhatt2019}
Samarth Brahmbhatt, Cusuh Ham, Charles~C. Kemp, and James Hays.
\newblock {ContactDB}: Analyzing and predicting grasp contact via thermal
  imaging.
\newblock In {\em The IEEE Conference on Computer Vision and Pattern
  Recognition ({CVPR})}. {IEEE}, jun 2019.

\bibitem{Brand2000}
Matthew Brand and Aaron Hertzmann.
\newblock Style machines.
\newblock In {\em Proceedings of the 27th Annual Conference on Computer
  Graphics and Interactive Techniques {(SIGGRAPH)}}. {ACM} Press, 2000.

\bibitem{caoHMP2020}
Zhe Cao, Hang Gao, Karttikeya Mangalam, Qi-Zhi Cai, Minh Vo, and Jitendra
  Malik.
\newblock Long-term human motion prediction with scene context.
\newblock In {\em European Conference on Computer Vision (ECCV)}. 2020.

\bibitem{Chao2017}
Yu-Wei Chao, Jimei Yang, Brian Price, Scott Cohen, and Jia Deng.
\newblock Forecasting human dynamics from static images.
\newblock In {\em {IEEE} Conference on Computer Vision and Pattern Recognition
  ({CVPR})}. {IEEE}, jul 2017.

\bibitem{Deng2021}
Fuqin Deng, Hua Feng, Mingjian Liang, Hongmin Wang, Yong Yang, Yuan Gao,
  Junfeng Chen, Junjie Hu, Xiyue Guo, and Tin~Lun Lam.
\newblock {FEANet}: Feature-enhanced attention network for {RGB}-thermal
  real-time semantic segmentation.
\newblock In {\em {IEEE}/{RSJ} International Conference on Intelligent Robots
  and Systems ({IROS})}. {IEEE}, sep 2021.

\bibitem{Fragkiadaki2015}
Katerina Fragkiadaki, Sergey Levine, Panna Felsen, and Jitendra Malik.
\newblock Recurrent network models for human dynamics.
\newblock In {\em {IEEE} International Conference on Computer Vision ({ICCV})}.
  {IEEE}, dec 2015.

\bibitem{Ha2017}
Qishen Ha, Kohei Watanabe, Takumi Karasawa, Yoshitaka Ushiku, and Tatsuya
  Harada.
\newblock {MFNet}: Towards real-time semantic segmentation for autonomous
  vehicles with multi-spectral scenes.
\newblock In {\em International Conference on Intelligent Robots and Systems
  ({IROS})}. {IEEE}, sep 2017.

\bibitem{He2016}
Kaiming He, Xiangyu Zhang, Shaoqing Ren, and Jian Sun.
\newblock Deep residual learning for image recognition.
\newblock In {\em {IEEE} Conference on Computer Vision and Pattern Recognition
  ({CVPR})}. {IEEE}, jun 2016.

\bibitem{Hernandez2019}
Alejandro Hernandez, Jurgen Gall, and Francesc Moreno.
\newblock Human motion prediction via spatio-temporal inpainting.
\newblock In {\em {IEEE} International Conference on Computer Vision ({ICCV})}.
  {IEEE}, oct 2019.

\bibitem{Huo2022}
Dong Huo, Jian Wang, Yiming Qian, and Yee-Hong Yang.
\newblock Glass segmentation with rgb-thermal image pairs.
\newblock Apr. 2022.

\bibitem{Ionescu2014}
Catalin Ionescu, Dragos Papava, Vlad Olaru, and Cristian Sminchisescu.
\newblock Human3.6m: Large scale datasets and predictive methods for 3d human
  sensing in natural environments.
\newblock {\em {IEEE} Transactions on Pattern Analysis and Machine Intelligence
  (TPAMI)}, 36(7):1325--1339, jul 2014.

\bibitem{Jain2016}
Ashesh Jain, Amir~R. Zamir, Silvio Savarese, and Ashutosh Saxena.
\newblock Structural-{RNN}: Deep learning on spatio-temporal graphs.
\newblock In {\em {IEEE} Conference on Computer Vision and Pattern Recognition
  ({CVPR})}. {IEEE}, jun 2016.

\bibitem{Jingchao2021}
Peng Jingchao, Zhao Haitao, Hu Zhengwei, Zhuang Yi, and Wang Bofan.
\newblock Siamese infrared and visible light fusion network for rgb-t tracking.
\newblock Mar. 2021.

\bibitem{Kristan2019}
Matej Kristan, Jiri Matas, Ales Leonardis, Michael Felsberg, Roman Pflugfelder,
  Joni-Kristian Kamarainen, Luka~Cehovin Zajc, Ondrej Drbohlav, Alan Lukezic,
  Amanda Berg, Abdelrahman Eldesokey, Jani Kapyla, Gustavo Fernandez, Abel
  Gonzalez-Garcia, Alireza Memarmoghadam, Andong Lu, Anfeng He, Anton
  Varfolomieiev, Antoni Chan, Ardhendu~Shekhar Tripathi, Arnold Smeulders,
  Bala~Suraj Pedasingu, Bao~Xin Chen, Baopeng Zhang, Baoyuan Wu, Bi Li, Bin He,
  Bin Yan, Bing Bai, Bing Li, Bo Li, Byeong~Hak Kim, Chao Ma, Chen Fang, Chen
  Qian, Cheng Chen, Chenglong Li, Chengquan Zhang, Chi-Yi Tsai, Chong Luo,
  Christian Micheloni, Chunhui Zhang, Dacheng Tao, Deepak Gupta, Dejia Song,
  Dong Wang, Efstratios Gavves, Eunu Yi, Fahad~Shahbaz Khan, Fangyi Zhang, Fei
  Wang, Fei Zhao, George~De Ath, Goutam Bhat, Guangqi Chen, Guangting Wang,
  Guoxuan Li, Hakan Cevikalp, Hao Du, Haojie Zhao, Hasan Saribas, Ho~Min Jung,
  Hongliang Bai, Hongyuan Yu, Hongyuan Yu, Houwen Peng, Huchuan Lu, Hui Li,
  Jiakun Li, Jianhua Li, Jianlong Fu, Jie Chen, Jie Gao, Jie Zhao, Jin Tang,
  Jing Li, Jingjing Wu, Jingtuo Liu, Jinqiao Wang, Jinqing Qi, Jinyue Zhang,
  John~K. Tsotsos, Jong~Hyuk Lee, Joost van~de Weijer, Josef Kittler, Jun~Ha
  Lee, Junfei Zhuang, Kangkai Zhang, Kangkang Wang, Kenan Dai, Lei Chen, Lei
  Liu, Leida Guo, Li Zhang, Liang Wang, Liangliang Wang, Lichao Zhang, Lijun
  Wang, Lijun Zhou, Linyu Zheng, Litu Rout, Luc~Van Gool, Luca Bertinetto,
  Martin Danelljan, Matteo Dunnhofer, Meng Ni, Min~Young Kim, Ming Tang,
  Ming-Hsuan Yang, Naveen Paluru, Niki Martinel, Pengfei Xu, Pengfei Zhang,
  Pengkun Zheng, Pengyu Zhang, Philip~H.S. Torr, Qi~Zhang~Qiang Wang, Qing Guo,
  Radu Timofte, Rama~Krishna Gorthi, Richard Everson, Ruize Han, Ruohan Zhang,
  Shan You, Shao-Chuan Zhao, Shengwei Zhao, Shihu Li, Shikun Li, Shiming Ge,
  Shuai Bai, Shuosen Guan, Tengfei Xing, Tianyang Xu, Tianyu Yang, Ting Zhang,
  Tomas Vojir, Wei Feng, Weiming Hu, Weizhao Wang, Wenjie Tang, Wenjun Zeng,
  Wenyu Liu, Xi Chen, Xi Qiu, Xiang Bai, Xiao-Jun Wu, Xiaoyun Yang, Xier Chen,
  Xin Li, Xing Sun, Xingyu Chen, Xinmei Tian, Xu Tang, Xue-Feng Zhu, Yan Huang,
  Yanan Chen, Yanchao Lian, Yang Gu, Yang Liu, Yanjie Chen, Yi Zhang, Yinda Xu,
  Yingming Wang, Yingping Li, Yu Zhou, Yuan Dong, Yufei Xu, Yunhua Zhang,
  Yunkun Li, Zeyu Wang~Zhao Luo, Zhaoliang Zhang, Zhen-Hua Feng, Zhenyu He,
  Zhichao Song, Zhihao Chen, Zhipeng Zhang, Zhirong Wu, Zhiwei Xiong, Zhongjian
  Huang, Zhu Teng, and Zihan Ni.
\newblock The seventh visual object tracking {VOT}2019 challenge results.
\newblock In {\em 2019 {IEEE}/{CVF} International Conference on Computer Vision
  Workshop ({ICCVW})}. {IEEE}, oct 2019.

\bibitem{Lan2021}
Xin Lan, Xiaojing Gu, and Xingsheng Gu.
\newblock {MMNet}: Multi-modal multi-stage network for {RGB}-t image semantic
  segmentation.
\newblock {\em Applied Intelligence}, 52(5):5817--5829, aug 2021.

\bibitem{Li2016}
Chenglong Li, Hui Cheng, Shiyi Hu, Xiaobai Liu, Jin Tang, and Liang Lin.
\newblock Learning collaborative sparse representation for grayscale-thermal
  tracking.
\newblock {\em {IEEE} Transactions on Image Processing}, 25(12):5743--5756, dec
  2016.

\bibitem{Li2019}
Chenglong Li, Xinyan Liang, Yijuan Lu, Nan Zhao, and Jin Tang.
\newblock {RGB}-t object tracking: Benchmark and baseline.
\newblock {\em Pattern Recognition}, 96:106977, dec 2019.

\bibitem{Li2022}
Chenglong Li, Wanlin Xue, Yaqing Jia, Zhichen Qu, Bin Luo, Jin Tang, and Dengdi
  Sun.
\newblock {LasHeR}: A large-scale high-diversity benchmark for {RGBT} tracking.
\newblock {\em {IEEE} Transactions on Image Processing}, 31:392--404, 2022.

\bibitem{Li2018}
Chen Li, Zhen Zhang, Wee~Sun Lee, and Gim~Hee Lee.
\newblock Convolutional sequence to sequence model for human dynamics.
\newblock In {\em {IEEE} Conference on Computer Vision and Pattern
  Recognition}. {IEEE}, jun 2018.

\bibitem{Li2017}
Chenglong Li, Nan Zhao, Yijuan Lu, Chengli Zhu, and Jin Tang.
\newblock Weighted sparse representation regularized graph learning for {RGB}-t
  object tracking.
\newblock In {\em Proceedings of the 25th {ACM} international conference on
  Multimedia}. {ACM}, oct 2017.

\bibitem{Li2019a}
Xueting Li, Sifei Liu, Kihwan Kim, Xiaolong Wang, Ming-Hsuan Yang, and Jan
  Kautz.
\newblock Putting humans in a scene: Learning affordance in 3d indoor
  environments.
\newblock In {\em {IEEE} Conference on Computer Vision and Pattern Recognition
  ({CVPR})}. {IEEE}, jun 2019.

\bibitem{Luo2019}
Chengwei Luo, Bin Sun, Ke Yang, Taoran Lu, and Wei-Chang Yeh.
\newblock Thermal infrared and visible sequences fusion tracking based on a
  hybrid tracking framework with adaptive weighting scheme.
\newblock {\em Infrared Physics and Technology}, 99:265--276, jun 2019.

\bibitem{Mao2019}
Wei Mao, Miaomiao Liu, Mathieu Salzmann, and Hongdong Li.
\newblock Learning trajectory dependencies for human motion prediction.
\newblock In {\em {IEEE} International Conference on Computer Vision ({ICCV})}.
  {IEEE}, oct 2019.

\bibitem{Martinez2017}
Julieta Martinez, Michael~J. Black, and Javier Romero.
\newblock On human motion prediction using recurrent neural networks.
\newblock In {\em {IEEE} Conference on Computer Vision and Pattern Recognition
  ({CVPR})}. {IEEE}, jul 2017.

\bibitem{Newell2016}
Alejandro Newell, Kaiyu Yang, and Jia Deng.
\newblock Stacked hourglass networks for human pose estimation.
\newblock In {\em European Conference on Computer Vision (ECCV)}, pages
  483--499. Springer International Publishing, 2016.

\bibitem{Shivakumar2020}
Shreyas~S. Shivakumar, Neil Rodrigues, Alex Zhou, Ian~D. Miller, Vijay Kumar,
  and Camillo~J. Taylor.
\newblock {PST}900: {RGB}-thermal calibration, dataset and segmentation
  network.
\newblock In {\em {IEEE} International Conference on Robotics and Automation
  ({ICRA})}. {IEEE}, may 2020.

\bibitem{Sun2019}
Yuxiang Sun, Weixun Zuo, and Ming Liu.
\newblock {RTFNet}: {RGB}-thermal fusion network for semantic segmentation of
  urban scenes.
\newblock {\em {IEEE} Robotics and Automation Letters}, 4(3):2576--2583, jul
  2019.

\bibitem{Sun2021}
Yuxiang Sun, Weixun Zuo, Peng Yun, Hengli Wang, and Ming Liu.
\newblock {FuseSeg}: Semantic segmentation of urban scenes based on {RGB} and
  thermal data fusion.
\newblock {\em {IEEE} Transactions on Automation Science and Engineering},
  18(3):1000--1011, jul 2021.

\bibitem{Vertens2020}
Johan Vertens, Jannik Zurn, and Wolfram Burgard.
\newblock {HeatNet}: Bridging the day-night domain gap in semantic segmentation
  with thermal images.
\newblock In {\em {IEEE}/{RSJ} International Conference on Intelligent Robots
  and Systems ({IROS})}. {IEEE}, oct 2020.

\bibitem{Walker2017}
Jacob Walker, Kenneth Marino, Abhinav Gupta, and Martial Hebert.
\newblock The pose knows: Video forecasting by generating pose futures.
\newblock In {\em {IEEE} International Conference on Computer Vision ({ICCV})}.
  {IEEE}, oct 2017.

\bibitem{Wang2020}
Chaoqun Wang, Chunyan Xu, Zhen Cui, Ling Zhou, Tong Zhang, Xiaoya Zhang, and
  Jian Yang.
\newblock Cross-modal pattern-propagation for {RGB}-t tracking.
\newblock In {\em 2020 {IEEE}/{CVF} Conference on Computer Vision and Pattern
  Recognition ({CVPR})}. {IEEE}, jun 2020.

\bibitem{Wang2021}
Jingbo Wang, Sijie Yan, Bo Dai, and Dahua Lin.
\newblock Scene-aware generative network for human motion synthesis.
\newblock In {\em {IEEE} Conference on Computer Vision and Pattern Recognition
  ({CVPR})}. {IEEE}, jun 2021.

\bibitem{Weng2019}
Chung-Yi Weng, Brian Curless, and Ira Kemelmacher-Shlizerman.
\newblock Photo wake-up: 3d character animation from a single photo.
\newblock In {\em {IEEE} Conference on Computer Vision and Pattern Recognition
  ({CVPR})}. {IEEE}, jun 2019.

\bibitem{Zhou2018}
Shuangjiu Xiao C. He Zeng Huang Hao~Li Yi~Zhou, Zimo~Li.
\newblock Auto-conditioned recurrent networks for extended complex human motion
  synthesis.
\newblock {\em International Conference on Learning Representation}, 2018.

\bibitem{Zhang2019a}
Jason Zhang, Panna Felsen, Angjoo Kanazawa, and Jitendra Malik.
\newblock Predicting 3d human dynamics from video.
\newblock In {\em {IEEE} International Conference on Computer Vision ({ICCV})}.
  {IEEE}, oct 2019.

\bibitem{Zhang2019}
Lichao Zhang, Martin Danelljan, Abel Gonzalez-Garcia, Joost van~de Weijer, and
  Fahad~Shahbaz Khan.
\newblock Multi-modal fusion for end-to-end {RGB}-t tracking.
\newblock In {\em 2019 {IEEE}/{CVF} International Conference on Computer Vision
  Workshop ({ICCVW})}. {IEEE}, oct 2019.

\bibitem{Zhang2021a}
Pengyu Zhang, Jie Zhao, Chunjuan Bo, Dong Wang, Huchuan Lu, and Xiaoyun Yang.
\newblock Jointly modeling motion and appearance cues for robust {RGB}-t
  tracking.
\newblock {\em {IEEE} Transactions on Image Processing}, 30:3335--3347, 2021.

\bibitem{Zhang2022}
Pengyu Zhang, Jie Zhao, Dong Wang, Huchuan Lu, and Xiang Ruan.
\newblock Visible-thermal uav tracking: A large-scale benchmark and new
  baseline.
\newblock Apr. 2022.

\bibitem{Zhang2021}
Qiang Zhang, Shenlu Zhao, Yongjiang Luo, Dingwen Zhang, Nianchang Huang, and
  Jungong Han.
\newblock {ABMDRNet}: Adaptive-weighted bi-directional modality difference
  reduction network for {RGB}-t semantic segmentation.
\newblock In {\em {IEEE}/{CVF} Conference on Computer Vision and Pattern
  Recognition ({CVPR})}. {IEEE}, jun 2021.

\bibitem{Zhang2022a}
Tianlu Zhang, Xueru Liu, Qiang Zhang, and Jungong Han.
\newblock {SiamCDA}: Complementarity- and distractor-aware {RGB}-t tracking
  based on siamese network.
\newblock {\em {IEEE} Transactions on Circuits and Systems for Video
  Technology}, 32(3):1403--1417, mar 2022.

\bibitem{Zhang2018}
Xingming Zhang, Xuehan Zhang, Xuedan Du, Xiangming Zhou, and Jun Yin.
\newblock Learning multi-domain convolutional network for {RGB}-t visual
  tracking.
\newblock In {\em 2018 11th International Congress on Image and Signal
  Processing, {BioMedical} Engineering and Informatics ({CISP}-{BMEI})}.
  {IEEE}, oct 2018.

\bibitem{Zhong2022}
Chongyang Zhong, Lei Hu, Zihao Zhang, Yongjing Ye, and Shihong Xia.
\newblock Spatio-temporal gating-adjacency gcn for human motion prediction.
\newblock {\em {IEEE} Conference on Computer Vision and Pattern Recognition
  ({CVPR})}, 2022.

\bibitem{Zhou2022}
Wujie Zhou, Shaohua Dong, Caie Xu, and Yaguan Qian.
\newblock Edge-aware guidance fusion network for {RGB}{\textendash}thermal
  scene parsing.
\newblock {\em Proceedings of the {AAAI} Conference on Artificial
  Intelligence}, 36(3):3571--3579, jun 2022.

\end{thebibliography}


\begin{thebibliography}{1}\itemsep=-1pt

\bibitem{easymocap}
Easymocap - make human motion capture easier.
\newblock Github, 2021.

\bibitem{openpose}
Z. {Cao}, G. {Hidalgo Martinez}, T. {Simon}, S. {Wei}, and Y.~A. {Sheikh}.
\newblock Openpose: Realtime multi-person 2d pose estimation using part
  affinity fields.
\newblock {\em IEEE Transactions on Pattern Analysis and Machine Intelligence},
  2019.

\bibitem{He2016}
Kaiming He, Xiangyu Zhang, Shaoqing Ren, and Jian Sun.
\newblock Deep residual learning for image recognition.
\newblock In {\em {IEEE} Conference on Computer Vision and Pattern Recognition
  ({CVPR})}. {IEEE}, jun 2016.

\bibitem{Joo2021}
Hanbyul Joo, Natalia Neverova, and Andrea Vedaldi.
\newblock Exemplar fine-tuning for 3d human model fitting towards in-the-wild
  3d human pose estimation.
\newblock In {\em International Conference on 3D Vision (3DV)}. {IEEE}.

\bibitem{Kingma2014}
Diederik~P. Kingma and Jimmy Ba.
\newblock Adam: A method for stochastic optimization.

\bibitem{SMPL:2015}
Matthew Loper, Naureen Mahmood, Javier Romero, Gerard Pons-Moll, and Michael~J.
  Black.
\newblock {SMPL}: A skinned multi-person linear model.
\newblock {\em ACM Trans. Graphics (Proc. SIGGRAPH Asia)}, 34(6):248:1--248:16,
  Oct. 2015.

\bibitem{Newell2016}
Alejandro Newell, Kaiyu Yang, and Jia Deng.
\newblock Stacked hourglass networks for human pose estimation.
\newblock In {\em European Conference on Computer Vision (ECCV)}, pages
  483--499. Springer International Publishing, 2016.

\end{thebibliography}
}

\end{document}


\title{Appendix}

\maketitle


\appendix

We first introduce the pose estimation pipeline in dataset construction in \cref{app:data}. In \cref{app:detail}, we specify the details of the method implementation. We individually evaluate each module in our method, and the results are in \cref{app:eval}. We also show some qualitative results in \cref{app:more} to demonstrate our method's capability of inferring the past. \cref{app:ablation} is an ablation study on our method pipeline. Finally, we state the approval for our Thermal-IM dataset in \cref{app:approval}.

\section{Pose estimation in dataset construction}
\label{app:data}
    We implement a two-stage pose extraction pipeline to acquire smooth and accurate 3D poses from RGB video pairs.

    In the first stage, we use \cite{Joo2021} to estimate a coarse 3D pose for every RGB frame. The resulting poses are in OpenPose\cite{openpose} \emph{body\_25} skeleton with 25 joints. According to the pose estimation results, we divide the videos into continuous segments, ensuring that the character is always in the view of both cameras in each segment.

    In the second stage, we synthesize the monocular pose estimation results of the two cameras to improve the pose quality. We first implement a triangulation step for more accurate depth estimation. Specifically, for each timestamp $t$, let $p_t,q_t$ be the coarse 3D poses in the camera coordinates detected from the two cameras. We find a pair of scales $a,b$ that minimizes the $\ell_2$ distance between $p_t \cdot a$ and $q_t \cdot b$ in the world coordinate. After that, we use EasyMocap \cite{easymocap} to refine the pose sequence further. It smooths a sequence of 3D poses by optimizing the SMPL body parameters \cite{SMPL:2015}.

    We further eliminate the failure cases of pose estimation. Specifically, all the poses are clustered into 2000 groups, and we manually filter out the clusters representing contorted poses.

\section{Implementation details}
\label{app:detail}
    \paragraph{Data processing:} In our task, we use the first 15 joints out of the 25 joints in the OpenPose skeleton to represent a human pose. The first 15 joints are enough to depict human actions while ignoring the details such as ears and toes.

    The input image size for our model is $288$ $\times$ $384$. And the evaluation metric MPJPE is also computed at this scale. The RGB and thermal lenses of our RGB-Thermal camera have different fields of view, and that of the thermal lens is smaller. We resize the thermal images in a preset way to align with the human poses, which are estimated in RGB image space.

    \paragraph{Model implementation:} The backbones of GoalNet and PoseNet are both an Hourglass model~\cite{Newell2016} with three blocks, while that of TypeNet is ResNet18~\cite{He2016}. The sizes of heatmap outputs of GoalNet and PoseNet are $72$ $\times$ $96$, and they are resized to be $288$ $\times$ $384$ by interpolation.

    All modules are trained using the Adam optimizer \cite{Kingma2014} for 6k batch iterations. The learning rates and batch sizes are in \cref{tab:hyper}. We use random crop and flip as data augmentation for all of them.

        \begin{table}
          \centering
          \begin{tabular}{ccc}
            \toprule
            Module & Learning rate & Batch size \\
            \midrule
            GoalNet & $5\times10^{-5}$ & $32$ \\	
            TypeNet & $5\times10^{-5}$ & $128$ \\
            PoseNet & $1\times10^{-4}$ & $32$ \\
            Semantic score model & $3\times10^{-5}$ & $128$ \\
            \bottomrule
          \end{tabular}
          \caption{\textbf{Learning rates and batch sizes.}}
          \label{tab:hyper}
        \end{table}

        \begin{figure}
            \centering
            \includegraphics[width=1\linewidth]{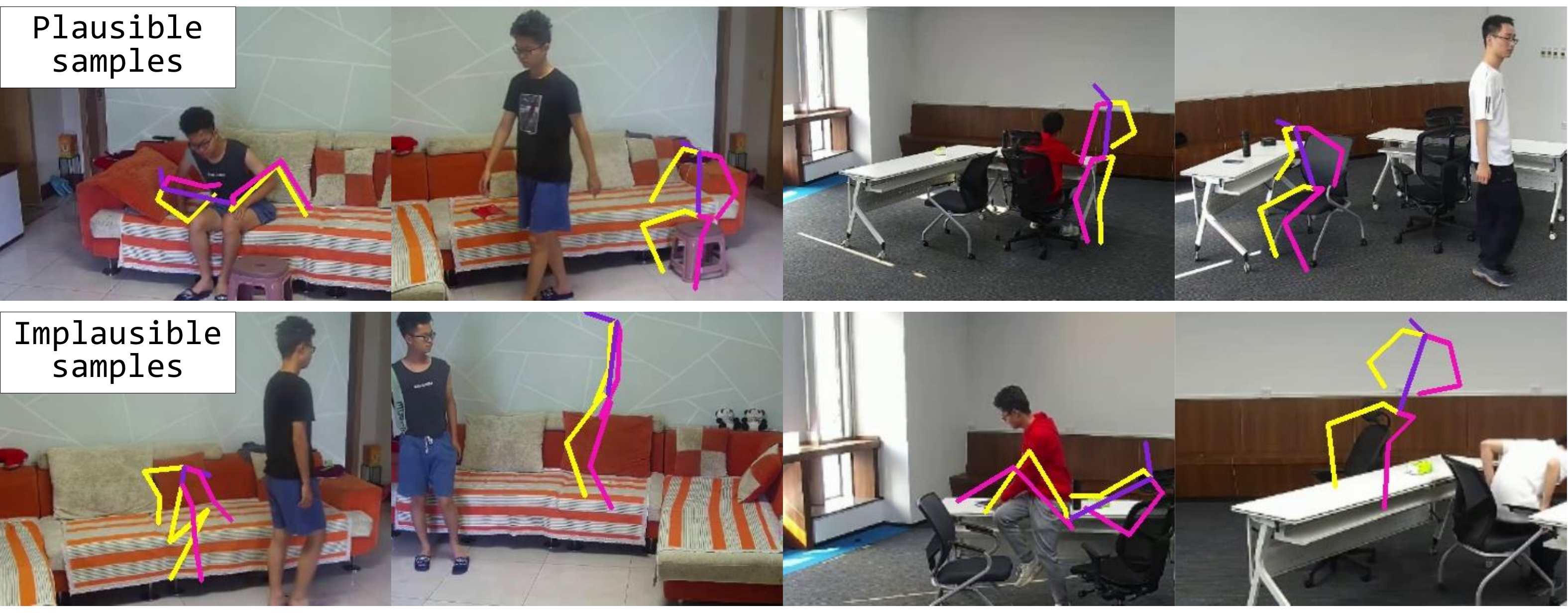}
            \caption{\textbf{Samples used to develop the semantic score classifier.} Plausible ones are samples in the dataset, while implausible ones are derived by random pose replacement, shift, and perturbation.}
            \label{fig:semantic}
        \end{figure}

    \paragraph{Semantic score:} The data we use to train the semantic score model contains RGB images with plausible and implausible poses. Plausible poses are the 3s-ago poses, and implausible poses are derived by randomly replacing, shifting, and perturbing the plausible ones. Some samples are shown in \cref{fig:semantic}.

    Given an RGB image and a pose, we want a binary classifier to estimate how likely the pose is plausible. We use ResNet18 as the model and train it with Binary Cross Entropy Loss. It is trained using the Adam optimizer with a weight decay of $1\times10^{-3}$ for 6k batch iterations. The learning rate and batch size are in \cref{tab:hyper}. Random crop and flip are used as data augmentation.

\section{Individual evaluation of modules}
\label{app:eval}

    As the three modules in our method are trained separately, we evaluate their performances in their own tasks in the following.

    \paragraph{GoalNet:} For each test instance, GoalNet samples 30 torso joint positions according to the predicted heatmap, and we evaluate how close they are to the ground truth 3s-ago position. We sort the 30 positions by order of their distances to the ground truth and compute the average $\ell_2$ distance of the top-1/3/5 ones. We show the results in \cref{tab:GoalNet}.

    \begin{table}
          \centering
          \begin{tabular}{cccc}
            \toprule
            \multirow{2}{*}{Module} & \multicolumn{3}{c}{Average $\ell_2$ Distance} \\
            \cmidrule(r){2-4}
            &  Top 1 & Top 3 & Top 5\\
            \midrule
             GoalNet & 10.50 & 15.02 & 31.12 \\
            \bottomrule
          \end{tabular}
          \caption{\textbf{Evaluation of GoalNet.} We calculate the $\ell_2$ distances from the top-1/3/5 predicted positions to the ground truth in the number of pixels.}
          \label{tab:GoalNet}
    \end{table}

    \paragraph{TypeNet:} We evaluate TypeNet as a classifier and report its top-1/3/5 accuracy. The results are in \cref{tab:TypeNet}.

    \begin{table}
          \centering
          \begin{tabular}{cccc}
            \toprule
            \multirow{2}{*}{Module} & \multicolumn{3}{c}{Accuracy} \\
            \cmidrule(r){2-4}
            &  Top 1 & Top 3 & Top 5\\
            \midrule
             TypeNet & 10.50 & 15.02 & 31.12 \\
            \bottomrule
          \end{tabular}
          \caption{\textbf{Evaluation of TypeNet.} The task of TypeNet is indeed classification, so we evaluate the top-1/3/5 accuracy of its prediction.}
          \label{tab:TypeNet}
    \end{table}

    \paragraph{PoseNet:} We examine how the refinement of PoseNet makes an inputted pose type center closer to the ground truth 3s-ago pose. We report the MPJPE of poses before and after refinement in \cref{tab:PoseNet}.

    \begin{table}
          \centering
          \begin{tabular}{ccc}
            \toprule
            \multirow{2}{*}{Module} & \multicolumn{2}{c}{MPJPE} \\
            \cmidrule(r){2-3}
            &  Before & After\\
            \midrule
             PoseNet & 8.87 & 8.59\\
            \bottomrule
          \end{tabular}
          \caption{\textbf{Evaluation of PoseNet.} Given the cluster center pose as input, we evaluate how much our PoseNet can refine it. The table shows the MPJPE from the poses to the ground truth poses before and after PoseNet refinement.}
          \label{tab:PoseNet}
    \end{table}

        \begin{table*}
          \centering
          \begin{tabular}{ccccccccc}
            \toprule
            \multicolumn{3}{c}{Modules} & & \multicolumn{3}{c}{MPJPE} & \multirow{2}{*}{NLL} & \multirow{2}{*}{Semantic Score($\%$)} \\
            \cmidrule(r){1-3}
            \cmidrule(r){5-7}
            GoalNet & TypeNet & PoseNet & &Top 1 & Top 3 & Top 5 & & \\
            \midrule
            \checkmark & & \checkmark  & & 19.04 & 22.45 & \textbf{25.19} & 112.28 & 73.12 \\
            \checkmark & \checkmark &  & & 18.64 & 22.61 & 25.65 & N/A & 76.68 \\
            \checkmark & \checkmark & \checkmark & & \textbf{18.33} & \textbf{22.25} & 25.25 & \textbf{103.75} & \textbf{82.11} \\
            \bottomrule
          \end{tabular}
          \caption{\textbf{Ablation study of removing different components.} Our model (the last row) outperforms the incomplete ones in most metrics, though removing TypeNet provides a slightly lower Top-5 MPJPE. We do not report the NLL for the one without PoseNet since it cannot be calculated in this setting.}
          \label{tab:ablation_module}
        \end{table*}

        \begin{figure}
            \centering
            \includegraphics[width=1\linewidth]{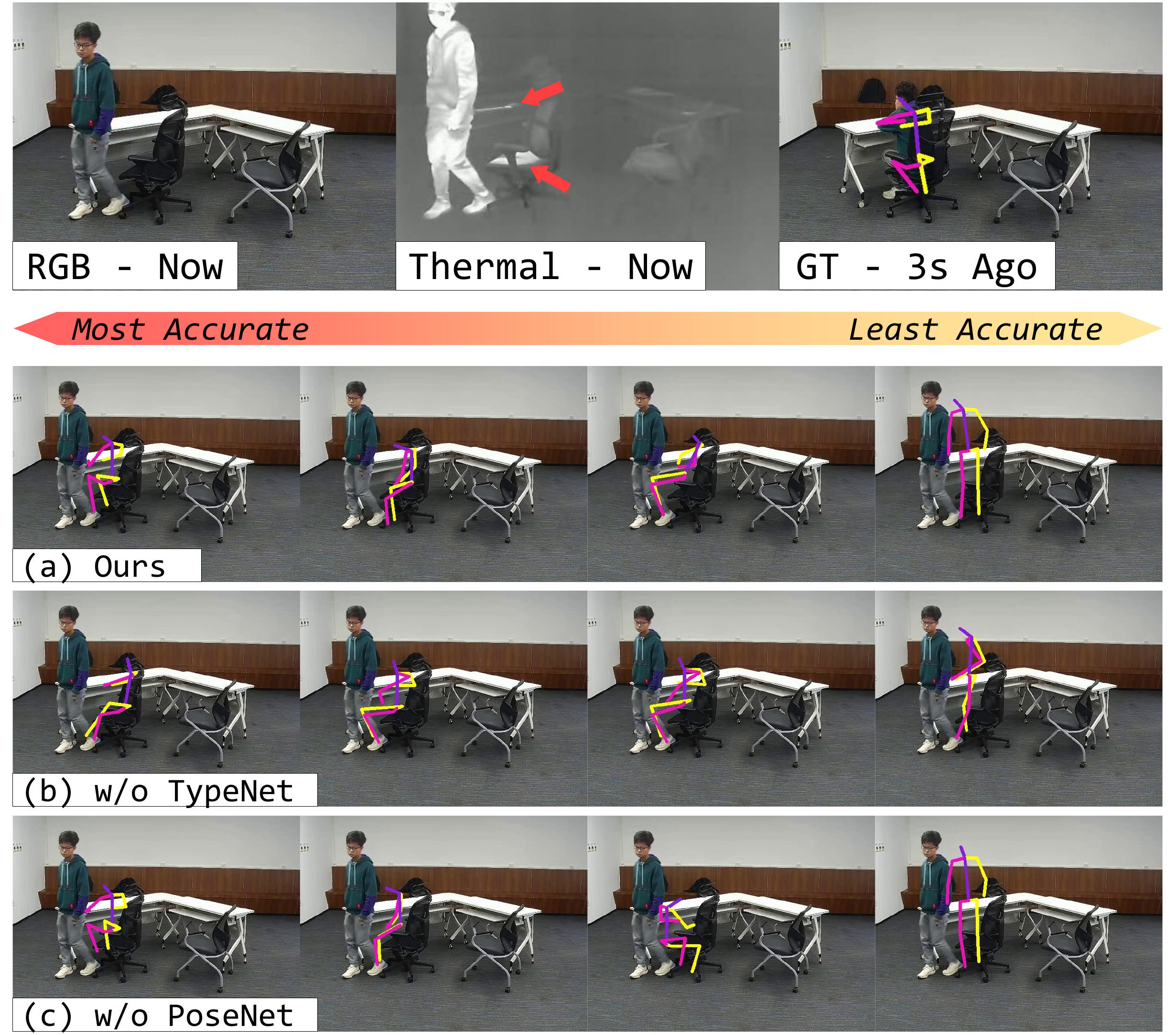}
            \caption{\textbf{Ablation study on model architecture.} From the thermal image, we can deduce that the person was sitting on the chair with arms on the table. In the model w/o TypeNet, predicted poses are often out-of-shape. The model w/o PoseNet can hardly provide the pose we desire because the number of pose types is limited. Our full model can refine the center pose of a type to fit with the details in the image, so it successfully generates sitting poses with an arm on the table (the 1st and 3rd column).}
            \label{fig:res3}
        \end{figure}

\section{More qualitative results}
\label{app:more}

\begin{figure*}
            \centering
            \includegraphics[width=1\linewidth]{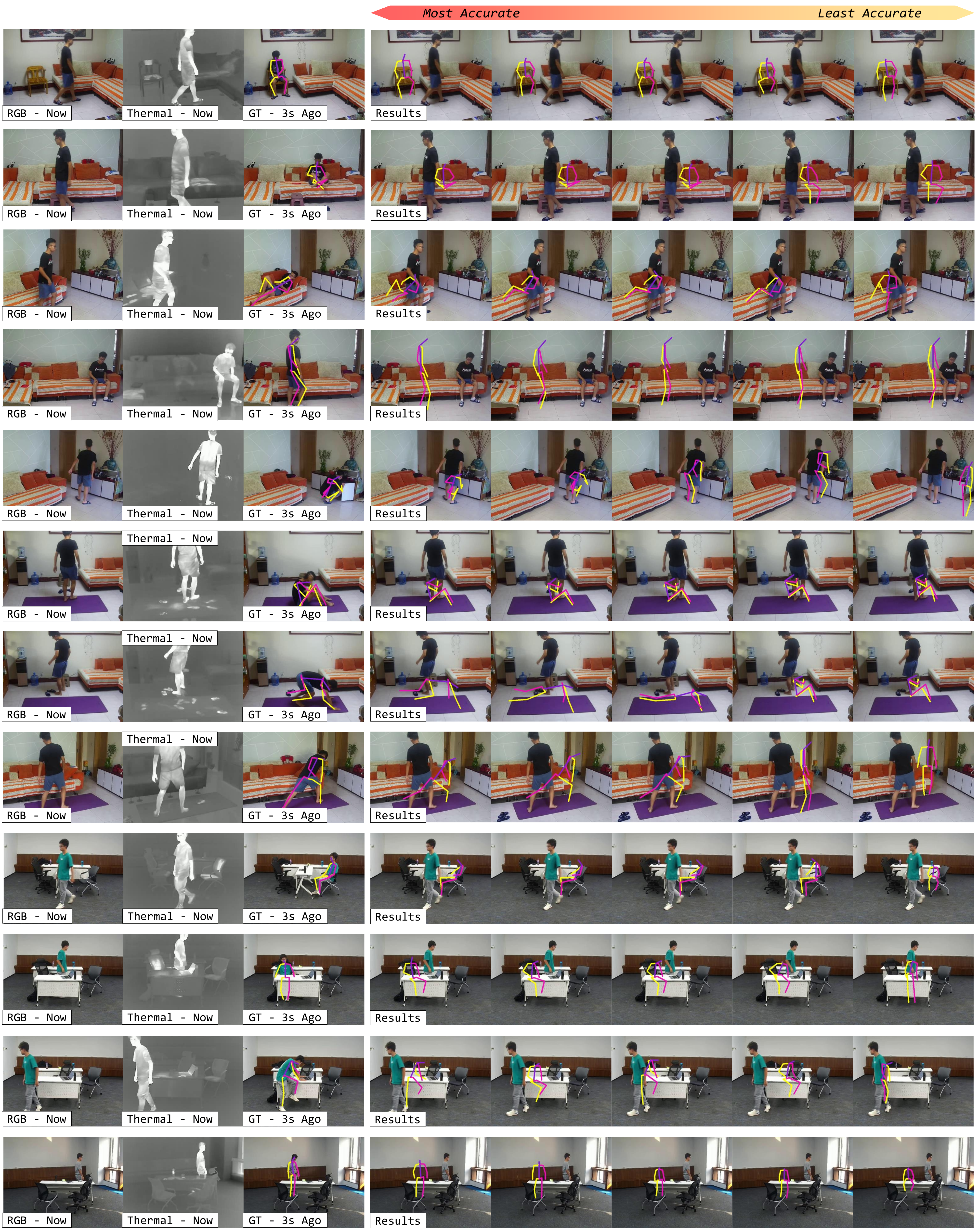}
            \caption{\textbf{Visualization results of our model.} For each sample, we sort the 30 predictions in order of MPJPE and show the 1st, 3rd, 5th, 10th, and 20th poses from left to right. Please pay attention to the bright marks pointed out by the arrows in the thermal images.}
            \label{fig:more}
        \end{figure*}

    In \cref{fig:more}, we show samples of our method's synthesized poses in the test set. The involved indoor actions include sitting on a sofa/chair/table, lying on a sofa, touching a cabinet/bottle, and several actions on a yoga mat (sit-ups, push-ups, and leg stretching).

\section{Ablation studies on pipeline modules}
\label{app:ablation}
    We implement two versions of our model without TypeNet or PoseNet to see how these modules contribute to our method.

    \paragraph{w/o TypeNet:} In a model without TypeNet, PoseNet generates a pose based on the input image and a root position given by GoalNet. The type of the synthesized pose is not specified here. In some cases, however, various poses are possible at a specific position. The skeleton joints generated by this model cannot be guaranteed to belong to the same pose, which leads to out-of-shape results as \cref{fig:res3}(b) shows and low semantic score in \cref{tab:ablation_module}. Besides, because the generated poses are far from reality, the top-1 MPJPE is much higher than our complete model, though the top-5 MPJPE is competitive.

    \paragraph{w/o PoseNet:} In a model without PoseNet, TypeNet provides a pose type, and the center pose of this type is moved to the GoalNet's predicted position to serve as an answer. Since the number of pose types is limited, the duplicated pose cannot always fit with the details in the image. In the first column of \cref{fig:res3}(a) \emph{vs.} (c), the model without PoseNet simply draws a sitting pose, while our complete model refines it so that the right arm is put on the table. As \cref{tab:ablation_module} illustrates, the refinement served by PoseNet improves both the synthesized poses' similarity to the answer and the plausibility in the context.

\section{Approval}
\label{app:approval}
    We have obtained approval for collecting and using the Thermal-IM dataset from the Institutional Review Board of our university department.

{\small
\bibliographystyle{ieee_fullname}
\bibliography{ref}
}